%% file: 0_main.tex
\documentclass[conference]{IEEEtran}
\IEEEoverridecommandlockouts
\usepackage{cite}
\usepackage{amsmath,amssymb,amsfonts}
\usepackage{algorithmic}
\usepackage{graphicx}
\usepackage{textcomp}
\usepackage{xcolor}

\usepackage{booktabs,makecell,multirow}
\usepackage{algorithm}

\usepackage{enumitem}

\def\BibTeX{{\rm B\kern-.05em{\sc i\kern-.025em b}\kern-.08em
    T\kern-.1667em\lower.7ex\hbox{E}\kern-.125emX}}

\makeatletter
\def\ps@IEEEtitlepagestyle{%
  \def\@oddfoot{\mycopyrightnotice}%
  \def\@evenfoot{}%
}
\def\mycopyrightnotice{%
  {\footnotesize\hfill This work has been accepted as an IEEE publication. Copyright © may be transferred without notice, after which this version may no longer be accessible.\hfill}%
  \gdef\mycopyrightnotice{}
}
    
\begin{document}

\title{Entailment-Driven Privacy Policy Classification with LLMs

\vspace{-3mm}
}

\author{
    \IEEEauthorblockN{
        Bhanuka Silva\textsuperscript{*}, 
        Dishanika Denipitiyage\textsuperscript{*}, 
        Suranga Seneviratne\textsuperscript{*}, 
        Anirban Mahanti\textsuperscript{*}, 
        Aruna Seneviratne\textsuperscript{†}
    }\\
    \vspace{-1mm}
    \IEEEauthorblockA{
        \textit{* School of Computer Science, The University of Sydney, Australia} \\
        \textit{† School of Electrical Engineering and Telecommunications, University of New South Wales, Australia} 
    }
}

\maketitle


\pagestyle{plain}

\begin{abstract}
While many online services provide privacy policies for end users to read and understand what personal data are being collected, these documents are often lengthy and complicated. As a result, the vast majority of users do not read them at all, leading to data collection under uninformed consent. Several attempts have been made to make privacy policies more user-friendly by summarising them, providing automatic annotations or labels for key sections, or by offering chat interfaces to ask specific questions. With recent advances in Large Language Models (LLMs), there is an opportunity to develop more effective tools to parse privacy policies and help users make informed decisions. In this paper, we propose an entailment-driven LLM-based framework to classify paragraphs of privacy policies into meaningful labels that are easily understood by users. The results demonstrate that our framework outperforms traditional LLM methods, improving the F1 score in average by 11.2\%. Additionally, our framework provides inherently explainable and meaningful predictions.

\end{abstract}


\input{1_introduction}
\input{2_related_work}

\input{3_solution}
\input{4_experiments}
\input{5_results}

\input{6_conclusion}

\bibliographystyle{IEEEtran}
\bibliography{references}    

\end{document}

%% file: 1_introduction.tex
\section{Introduction}
\label{Sec:Introduction}

Many online services and apps we use today collect vast volumes of personal data~\cite{binns2018third}. Beyond the first-party use of this data for purposes such as personalisation, it is often used for advertising, analytics, and user profiling. Additionally, this data can be shared with, or even sold to, third parties without the direct knowledge of users, posing serious privacy risks~\cite{binns2018third, habib2019empirical}. Typically, information regarding such data collection and sharing practices is outlined in the service's privacy policies and providing the users with privacy policies is mandatory in many jurisdictions~\cite{GDPRthirdParty}. However, these policies are usually lengthy, complicated, and written in complex legal jargon.
As a result, users frequently agree to data collection practices without thoroughly reading the privacy policies or comprehending the potential risks involved.

While some service providers actively attempt to enhance the readability of privacy policies~\cite{Google_privacy_policy,krumay2020readability}, the vast majority of these documents remain incomprehensible to end-users. Consequently, multiple research efforts have explored the possibility of providing users with more user-friendly representations of complex privacy policies. These approaches range from presenting user-friendly labels~\cite{privacy_labels_2015}, to designing chatbots to answer privacy-related questions~\cite{harkous2018}. However, most of the existing work has leveraged classical Natural Language Processing (NLP) techniques and encoder-only language models, such as variants of the BERT model.

Recent advancements in Large Language Models (LLMs), such as GPT~\cite{gpt2, gpt4} and LLaMA~\cite{llama2}, have established them as the state-of-the-art for a majority of NLP tasks. These models have demonstrated excellent capabilities in areas like text summarisation and understanding. Moreover, the versatility of LLMs has been showcased in various application domains, including medicine~\cite{waisberg2023gpt}, finance~\cite{wu2023bloomberggpt}, and others. These developments in LLMs provide a foundation for building novel solutions that can extract useful information from otherwise complex and nearly unreadable privacy policies, and present it to users in a more user-friendly manner.

In this paper, we propose a novel entailment-driven LLM-based framework for privacy policy paragraph classification. Traditional LLM approaches are known to suffer from well-known hallucination problems and thus may not always generate the expected result directly (e.g., summarised text, classification label of text). As a result, the onus of determining whether or not the LLM has made up or dropped facts will be on the users. One key idea behind our approach is to bolster LLM-based classification frameworks with an additional “entailment” phase to filter out the initial classifications by LLMs in an analogous way to how we would select or drop a particular LLM-generated output.
Fig.~\ref{fig:intro_example} demonstrates this phenomenon with an example. At stage 1, an \emph{explained classifier} predicts a class output and a corresponding reason for a given privacy text. Then, we mask the reason from the original text and use an intermediate stage 2 with a \emph{blank filler} in an attempt to predict the reasoning text again. An \emph{entailment verifier} receives information from stages 1 and 2 both and decides `entailment', i.e., the class prediction and the original reasoning are acceptable or vice versa.
More specifically,

\begin{itemize}[leftmargin=*]

    \item We propose an entailment-driven LLM-based framework to classify paragraphs in privacy policies into 12 categories, such as \emph{first party collection/use}, \emph{third party sharing/collection}, and \emph{user choice/control}, that are easier to understand by the users. The key components of our framework include the explained classifier that generates classification thoughts, the blank filler that re-thinks about these original thoughts and the entailment verifier that makes the final decision (analogous to how a human would reason).
    
    \item We evaluate the performance of our proposed method using the OPP-115 dataset and compare our results with existing baselines and zero-shot LLM settings. We find that our method performs better than vanilla LLM-methods; 8.6\%, 14.5\%, and 10.5\% higher than the results of T5, GPT4, and LLaMA2, respectively in terms of macro-average F1 score.

    \item We further analyse the explainability of our method and show quantitatively that it is better than other baseline methods we compare with. Out of 57.9\% of the predictions, our method generates reasoning texts that are at least 50\% or more overlapping with what a legal-expert would have reasoned. Comparatively, it is only 18.3\% for the best performing embedding-based model of PrivBERT.
\end{itemize}

The rest of the paper is organised as follows: Section ~\ref{Sec:RelatedWork} discusses related work. Section~\ref{Sec:Framework} details our framework, while Section~\ref{sec: experimental setup} describes the experiment setup. Results and comparisons with other baselines are provided in Section~\ref{Sec: Results}. Finally, Section~\ref{sec:conclusion} concludes the paper.

%% file: 2_related_work.tex
\section{Related Work}
\label{Sec:RelatedWork}

\subsection{Empirical Studies on Privacy Policies}

Users are increasingly concerned about online privacy~\cite{consumers_aware}, yet empirical studies consistently show that privacy policy documents have become substantially longer over the past two decades. With median word counts ranging from 1,500~\cite{www21_PP_overtime} to 2,500~\cite{PP_apps_for_youth}, these documents take too long to read~\cite{2008_cost_of_reading, www21_PP_overtime}, resulting in users making little effort to read and understand them~\cite{why_ignore_pp}. Further, the writing and presentation of these documents often make them inaccessible~\cite{ harkous2018}, with the end result often being uninformed consent~\cite{gdpr}.

Recent regulatory and compliance attempts, such as the EU GDPR, have aimed to make privacy policies mandatory and more user-friendly. While these efforts have led to positive outcomes like a 4.9\% increase in the availability of privacy policies~\cite{take_some_cookies}, a longitudinal analysis by~\cite{policy_landscape} finds that this has caused privacy policies to become even longer, increasing by around 25\% globally. This makes the policies even more challenging to read and therefore, to understand.

\subsection{Analysing Privacy Policies Using NLP Techniques}

Relying solely on manual analysis of privacy policies or manually crafted rules, as in~\cite{policylint, bhatia2016automated}, does not scale. As a result, many researchers have proposed automated methods to provide end-users with meaningful interpretations of privacy documents. Some examples include identifying relevant privacy topics for policy sections~\cite{liu2018towards, privbert}, summarising sections~\cite{gopinath2020automatic}, highlighting subsections where users can make informed choices regarding their personal information (e.g., opt-out choices~\cite{sathyendra2016, sathyendra2017identifying}), developing question-answering systems for policy documents~\cite{harkous2018}, building user-interface tools to identify common data practices (e.g., browser extensions~\cite{zimmeck2014privee}), or detecting privacy policy inconsistencies and non-compliances~\cite{story2019natural, policy_landscape}. The majority of these works have leveraged traditional Natural Language Processing (NLP) techniques together with machine learning methods such as support vector classifiers, logistic regression models, and neural networks, with limited adoption of recent transformer-based language encoder models like BERT~\cite{bert}. PrivBERT~\cite{privbert} is a domain adapted version of a popular encoder model RoBERTa~\cite{roberta} that performs better than vanilla-encoder models in classification tasks. 

\begin{figure}[t]
    \centering
    \includegraphics[width=0.45\textwidth]{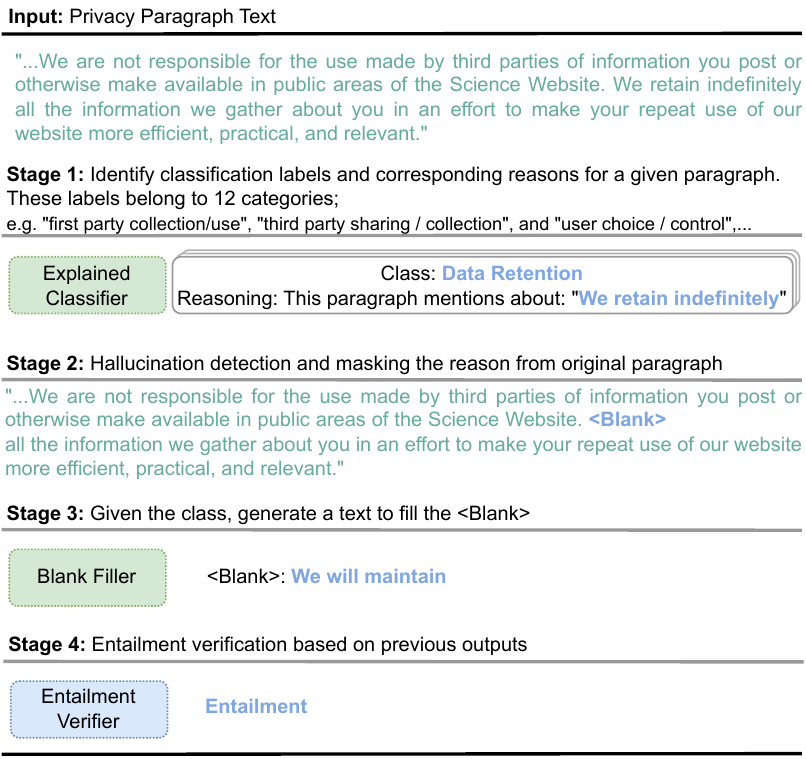}
    \caption{Four stages of the entailment-driven privacy policy classification. Depicted paragraph is from our test-dataset and all of the outputs generated at stage 1 would undergo stage 2 to 4 separately.}
    \label{fig:intro_example}
    \vspace{-3ex}
\end{figure}


\subsection{Large Language Models (LLMs)}

The recent surge of Large Language models (LLMs) such as GPT4~\cite{gpt4}, LLaMA2~\cite{llama2} has resulted in generative AI models being adapted with a wide range of tasks, including world knowledge, commonsense, and summarisation. The performance of LLMs could be further enhanced through domain adaptation, as demonstrated by initiatives like BloombergGPT~\cite{wu2023bloomberggpt} for the financial sector and Code LLaMA~\cite{roziere2023code} for software development. However, such endeavors are resource-intensive, requiring substantial text data and computational power.

PolicyGPT~\cite{tang2023policygpt} is a recent attempt to explore zero-shot prompting for privacy policy paragraph classification with LLMs. 
However, our experiments show that PolicyGPT's zero-shot performance falls short when confronted with multi-class-multi-label classification (refer Sec.~\ref{sec: evaluation metrics}).  

In this work, leveraging open-source LLMs,  we explore how LLMs' unique explainable capabilities can aid in better interpretations of privacy policies. State-of-the-art chain-of-thought (CoT) prompting~\cite{cot}, which maps non-trivial inputs and outputs via intermediate steps, mimics the human thought process by breaking down a complex task into smaller, more interpretable steps. Drawing inspiration from CoT prompting, we investigate how the inclusion of one-step reasoning; \emph{i.e. a `reason' shown in stage 1 of Fig.~\ref{fig:intro_example}}, can improve our classification accuracy and lead to more reliable and explainable outputs.
\vspace{-2mm}

%% file: 3_solution.tex
\section{Our Framework}
\label{Sec:Framework}

\begin{figure*}[ht]
    \centering
    \includegraphics[width=0.7\textwidth]{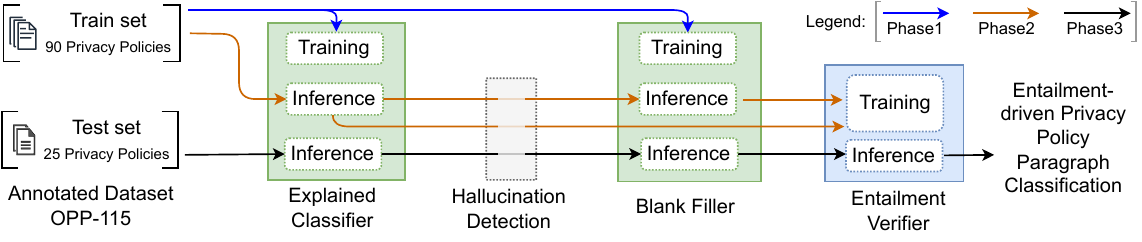}
    \caption{End to end pipeline of our method. Shaded in green are decoder models and shaded in blue is an encoder model. Phase 1 represents the training of explained classifier and blank filler. Phase 2 represents the training of the entailment verifier, for which we use the already trained modules from phase 1 in inference mode. Phase 3 represents all three modules working in inference mode with test dataset. }
    \label{fig:pipeline}
    \vspace{-3ex}
\end{figure*}

Our framework starts with explainable privacy policy paragraph classification. This is depicted in the first block of Fig.~\ref{fig:pipeline} named as explained classifier. Next, an explanation-masked-out version of the privacy paragraph and previous classification output are given as inputs to the blank filler. It will then generate the most likely token sequence for that `masked portion' by looking at the broader context of the paragraph. Finally, we check the entailment among these via the last block of our pipeline, entailment verifier. How each module works is explained below in detail.

\subsection{Explained Classifier}
The explained classifier's intuition is to generate a sufficient number of class identifications with reasons that are extracted from a privacy policy paragraph. Therefore, this module requires an autoregressive LLM. Formally; an output class label $y_i \in [c_0, c_1, .. c_n]$ will be predicted alongside a subset of original tokenised text, i.e., a reason $t_i \in T$ during its training and inference stages. Here, $c_0...c_n$ are all the class labels defined in the annotated dataset. $T$ is the tokenised privacy policy paragraph text chunk that the model is fed with. $t_i \in T$ condition is evaluated using Python Regex based hallucination detector and such filtered $y_i, t_i$ pairs will be forwarded to the next stage of the model. For each paragraph, there could be any number of such pairs establishing a multi-label setting. 

\subsection{Blank Filler}
Blank filler is also an autoregressive LLM that takes paragraph $T$ as input where a reason $t_i$ from the explained classifier is masked out and would try to predict the best text chunk $t'_i$ to fill in that masked part of the paragraph. We further provide the explained classifier's output class label $y_i$ as an input indicating the model which kind of text it should try to regenerate. We explore the model’s understanding of the paragraph’s general landscape here; ``if we mask out the main reason for a particular classification output, can the model look at the rest of the paragraph and then predict what kind of text should be there?". It is worthwhile to emphasise that we do not expect the outputs $t'_i$ to be word-for-word identical to $t_i$. Instead, we expect $t'_i$ to be similar in meaning to $t_i$. 

\subsection{Entailment Verifier}
\label{sec: entailment verifier}
This is an encoder-based language model attached to a neural network classifier head where we feed previously generated outputs $y_i$, $t_i$, and $t'_i$ separated by $[SEP]$ tokens. The output of this module is binary, indicating entailment (output:1) or contradiction (output:0). This module acts as the final filter where we can remove contradictory classifications and reasons from prediction outputs for a given policy paragraph $T$. 

We further explain how we train each of these modules using the training dataset and how we perform entitlement-driven privacy policy paragraph classifications for the testing dataset in more detail in Sec.~\ref{sec: pipeline}. 

%% file: 4_experiments.tex
\section{Experimental Setup}
\label{sec: experimental setup}

This section provides an overview of the experimental setup, including a description of the annotated dataset we use and the train-test split we selected. It then introduces the seven baseline models against which we compare our proposed framework, followed by an outline of the evaluation criteria.

\subsection{Modules of the Framework}

We used two 8-bit quantised LLaMA2 models~\cite{llama2} with low-rank adaptation (LoRA)~\cite{lora} for our explained classifier and blank filler modules. Although our framework can accommodate any large language model, we chose LLaMA2 due to its open-source availability and its superior performance in tasks such as commonsense reasoning, world knowledge, and reading comprehension compared to the state-of-the-art at the time of its release.

For the entailment verifier, we selected a BERT encoder module with 110 million parameters that demonstrated good results for the Multi-Genre Natural Language Inference (MNLI), which is also an entailment classification task~\cite{bert}. Again, we emphasise that our framework is flexible enough to incorporate any other encoder model as the entailment verifier. We fine-tune the BERT model based on the inference results from the explained classifier and blank filler ({\bf cf.} Sec.~\ref{sec: pipeline}).

\subsection{OPP-115 Dataset}

\begin{figure}
    \centering
    \includegraphics[width = 0.48\textwidth]{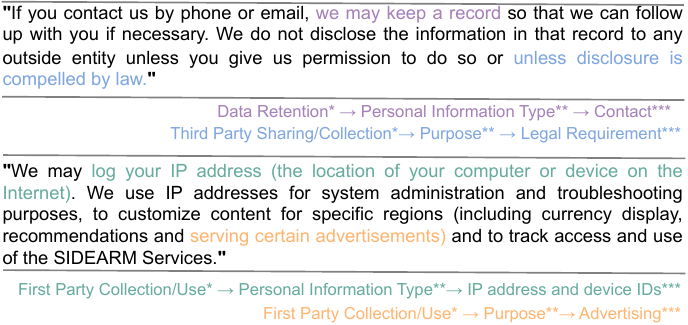}
    \caption{Two example paragraphs from train-set depicting the annotated segment and the relevant, data-practice label*, data-attribute label**, and data-value label***}
    \label{fig:dataset-examples}
    \vspace{-3ex}
\end{figure}

To evaluate the performance of our framework, we used the OPP-115 dataset~\cite{opp115}, which contains paragraph excerpts of online privacy policies annotated and labelled by legal experts. It is commonly used in comparable work~\cite{sathyendra2016, story2019natural, harkous2018, privbert}.
Each paragraph excerpt in the dataset (extracted from 115 web privacy policies) has a 3-tiered label. The highest level of the labelling tier is called ``data practice" (e.g., \emph{first party collection/use}), and there are ten such labels. Each high-level tier subsequently has fine-granular labels according to ``data-attribute" (e.g., \emph{personal information type})  and ``data-value" (e.g., \emph{contact}). Additionally, the paragraph excerpt also has the corresponding parts that led to specific labelling annotated. It is important to note here that one paragraph can have multiple places annotated with different labels. We show two example paragraphs from the dataset in Fig.~\ref{fig:dataset-examples}. 

Comparable to other work~\cite{tang2023policygpt,privbert}, the main task we solve in this dataset is given a paragraph, predicting the correct first-tier label(s). One of the ten data practice classes called \emph{Other}, is ambiguous in that a model can not give a reason unless given a definition of what \emph{other} means. Therefore, we augment it with the information from a lower-level ``data attribute''. This transformation makes our final problem a 12-class classification problem.  This is consistent with how other baseline methods, such as PrivBERT~\cite{privbert}, used the OPP-115 dataset. Here, we highlight that the exact problem is a multi-class (out of 12 classes) and multi-label setting since one paragraph can contain multiple annotations with different corresponding labels. For the train-test split, out of the 115 privacy policies in the dataset, we used 2948 paragraphs associated with 90 privacy policies for training and the 683 paragraphs associated with the remaining 25 privacy policies for testing. In the training set, 54.5\% paragraphs contain only 1 label, 32.1\% contain 2 labels, and the remaining with 3 or more labels. Test set consists a similar distribution as well. 
\vspace{-6mm}

\subsection{Training/Testing Pipeline}
\label{sec: pipeline}

Initially, we train the explained classifier and blank filler separately with the training data in supervised fine-tuning setting over five epochs (Phase1 in Fig.~\ref{fig:pipeline}). Next, we create a labelled dataset for training the entailment verifier, according to the inputs outlined in Sec.~\ref{sec: entailment verifier} and illustrated in Phase2 of Fig.~\ref{fig:pipeline}. This dataset is created by running the previously fine-tuned explained classifier and blank-filler in inference mode over the training set.
During this inference phase, the explained classifier generates incorrect class predictions that did not exist in the training-set. We denote these made up classes as `contradictions'. Conversely, any accurate class prediction the explained classifier makes is marked as an `entailment'. This approach removes the necessity of manually augmenting and curating another dataset, just for training the entailment verifier. Further, it enables us to expose the entailment verifier to realistic errors made by the previous modules, thereby facilitating its training to recognise such mistakes.
Once all three modules are trained, we deploy them in inference mode, as shown in Phase3 of Fig.~\ref{fig:pipeline}, to evaluate performance on the held-out test set.

\subsection{Baselines}

We compare the performance of our method against seven baselines falling under two broad categories. 

\subsubsection{Embedding-Based Classification Models}

These models have demonstrated effectiveness in text-classification tasks~\cite{bert, roberta}. Typically, embeddings extracted from a language model are processed by a linear classification head to generate the final predictions. For baselines in this category, we employ two generic encoder models, BERT and RoBERTa, then GPT2, which is adapted to the classification task by using the mean embedding representation of the token sequence of a given text input and finally, PrivBERT, which was further pre-trained on privacy policies for domain adaptation.

\subsubsection{Language Generation Based Classification Models}

The next set of baselines are auto-regressive language generation-based classifiers. In other words, these models generate the most likely classification outputs in the form of natural text. These baselines closely resemble our method as our explained classifier operates in this setting. We evaluate the performance of LLaMA2 in a supervised fine-tuning setting as the vanilla LLM, where the model outputs the class labels in natural text. Next, we adapt the encoder-decoder T5~\cite{t5} model by feeding policy text in the training set as inputs and fine-tuning it to generate target text that represents the class labels. Finally, we adopt GPT4 as a baseline model using a prompting structure introduced by~\cite{tang2023policygpt} to suit the twelve-class, multi-label setting and perform a zero-shot evaluation for our test dataset.

\subsection{Evaluation Metrics}
\label{sec: evaluation metrics}

\begin{table*}
    \centering
    \caption{Performance Comparison} \vspace{-2mm}
    \label{tab: baselines}

    \begin{tabular}{l|ccc|ccc|ccc|ccc|r }
        \multicolumn{14}{l}{\textbf{(a) Embedding based classification models}}\\
        
        \hline
        \vspace{-2ex} 
        & \multicolumn{3}{c|}{} 
        & \multicolumn{3}{c|}{} 
        & \multicolumn{3}{c|}{} 
        & \multicolumn{3}{c|}{} &  \\ 
        
             & \multicolumn{3}{c|}{GPT2 Embedding} 
             & \multicolumn{3}{c|}{BERT} 
             & \multicolumn{3}{c|}{RoBERTa} 
             & \multicolumn{3}{c|}{PrivBERT} &  \\
        Class                               & P&R&F1             & P&R&F1             & P&R&F1             & P&R&F1             & Support  \\
        
        \hline
        \vspace{-1ex} 
        & \multicolumn{3}{c|}{} 
        & \multicolumn{3}{c|}{} 
        & \multicolumn{3}{c|}{} 
        & \multicolumn{3}{c|}{} &  \\ 
        First Party Collection/Use          &0.87 & 0.77 & 0.82  &0.90 & 0.80 & 0.85  &0.84 & 0.88 & 0.86  &0.88 & 0.85 & 0.87  & 289\\
        Third Party Sharing/Collection      &0.84 & 0.83 & 0.84  &0.88 & 0.85 & 0.87  &0.83 & 0.86 & 0.85  &0.94 & 0.83 & 0.88  &204\\
        User Choice/Control                 &0.93 & 0.48 & 0.64  &0.86 & 0.54 & 0.66  &0.77 & 0.58 & 0.66  &0.82 & 0.69 & 0.75  &115\\
        User Access, Edit and Deletion      &0.85 & 0.80 & 0.86  &0.74 & 0.85 & 0.80  &0.74 & 0.78 & 0.76  &0.78 & 0.76 & 0.77  &41\\
        Introductory/Generic                &0.75 & 0.56 & 0.64  &0.69 & 0.53 & 0.60  &0.76 & 0.44 & 0.56  &0.73 & 0.69 & 0.71  &118\\
        Policy Change                       &0.80 & 0.55 & 0.65  &0.83 & 0.52 & 0.64  &0.78 & 0.48 & 0.60  &0.80 & 0.55 & 0.65  &29\\
        Data Security                       &0.93 & 0.61 & 0.74  &0.79 & 0.71 & 0.75  &0.73 & 0.73 & 0.73  &0.70 & 0.76 & 0.73  &62\\
        International \& Specific Audience  &0.90 & 0.78 & 0.84  &0.84 & 0.85 & 0.84  &0.83 & 0.87 & 0.85  &0.88 & 0.88 & 0.88  &60\\
        Practice Not Covered                &0.61 & 0.35 & 0.44  &0.61 & 0.42 & 0.50  &0.62 & 0.32 & 0.42  &0.67 & 0.34 & 0.45  &110\\
        Data Retention                      &0.56 & 0.58 & 0.57  &1.00 & 0.35 & 0.51  &0.67 & 0.46 & 0.55  &0.94 & 0.65 & 0.77  &26\\
        Privacy Contact Information         &0.93 & 0.68 & 0.78  &0.85 & 0.80 & 0.83  &0.89 & 0.88 & 0.88  &0.82 & 0.89 & 0.85  &56\\
        Do Not Track                        &1.00 & 0.60 & 0.75  &1.00 & 0.20 & 0.33  &1.00 & 0.80 & 0.89  &1.00 & 0.60 & 0.75  &5\\
        \hline
        \vspace{-1ex} 
        & \multicolumn{3}{c|}{} 
        & \multicolumn{3}{c|}{} 
        & \multicolumn{3}{c|}{} 
        & \multicolumn{3}{c|}{} &  \\ 
        Micro Average                       &0.84 & 0.66 & 0.74  &0.83 & 0.70 & 0.76 &0.80 & 0.71 & 0.75  &0.84 & 0.74 & {\bf 0.79}  &1115\\
        Macro Average                       &0.83 & 0.63 & 0.71  &0.83 & 0.62 & 0.68 &0.79 & 0.67 & 0.72  &0.83 & 0.71 & {\bf 0.76}  &1115\\
        Weighted Average                    &0.83 & 0.66 & 0.73  &0.83 & 0.70 & 0.75 &0.79 & 0.71 & 0.74  &0.83 & 0.74 & {\bf 0.78}  &1115\\
        \hline
        \multicolumn{14}{l}{}\\
        \multicolumn{14}{l}{\textbf{(b) Language generation based classification models}}\\
        \hline
        \vspace{-2ex} 
        & \multicolumn{3}{c|}{} 
        & \multicolumn{3}{c|}{} 
        & \multicolumn{3}{c|}{} 
        & \multicolumn{3}{c|}{} &  \\ 
             & \multicolumn{3}{c|}{T5} 
             & \multicolumn{3}{c|}{GPT4 Prompting} 
             & \multicolumn{3}{c|}{LLaMA2 7B Fine-tuned} 
             & \multicolumn{3}{c|}{Our Method} &  \\
        Class                               & P&R&F1             & P&R&F1             & P&R&F1             & P&R&F1            & Support  \\
        \hline
        \vspace{-1ex} 
        & \multicolumn{3}{c|}{} 
        & \multicolumn{3}{c|}{} 
        & \multicolumn{3}{c|}{} 
        & \multicolumn{3}{c|}{} &  \\ 
        First Party Collection/Use          &0.86 & 0.78 & 0.82  &0.68 & 0.91 & 0.78  &0.92 & 0.66 & 0.77  &0.83 & 0.72 & 0.77  &289 \\
        Third Party Sharing/Collection      &0.87 & 0.65 & 0.74  &0.70 & 0.76 & 0.73  &0.92 & 0.48 & 0.63  &0.82 & 0.78 & 0.80  &204 \\
        User Choice/Control                 &0.75 & 0.43 & 0.55  &0.47 & 0.71 & 0.57  &0.87 & 0.39 & 0.54  &0.62 & 0.67 & 0.64  &115 \\
        User Access, Edit and Deletion      &0.86 & 0.61 & 0.71  &0.47 & 0.80 & 0.59  &0.83 & 0.49 & 0.62  &0.67 & 0.70 & 0.68  &41  \\
        Introductory/Generic                &0.72 & 0.43 & 0.54  &0.69 & 0.31 & 0.43  &0.43 & 0.75 & 0.55  &0.56 & 0.44 & 0.50  &118 \\
        Policy Change                       &0.93 & 0.48 & 0.64  &0.43 & 0.79 & 0.55  &1.00 & 0.48 & 0.65  &0.54 & 0.56 & 0.55  &29  \\
        Data Security                       &0.89 & 0.52 & 0.65  &0.52 & 0.77 & 0.62  &0.94 & 0.48 & 0.64  &0.83 & 0.69 & 0.75  &62  \\
        International \& Specific Audience  &0.86 & 0.70 & 0.77  &0.64 & 0.87 & 0.74  &0.73 & 0.72 & 0.72  &0.64 & 0.67 & 0.66  &60  \\
        Practice Not Covered                &0.27 & 0.03 & 0.05  &0.44 & 0.25 & 0.32  &0.33 & 0.01 & 0.02  &0.48 & 0.33 & 0.39  &110 \\
        Data Retention                      &0.75 & 0.12 & 0.20  &0.38 & 0.46 & 0.41  &1.00 & 0.19 & 0.32  &0.59 & 0.50 & 0.54  &26  \\
        Privacy Contact Information         &1.00 & 0.39 & 0.56  &0.49 & 0.75 & 0.59  &1.00 & 0.41 & 0.58  &0.73 & 0.75 & 0.74  &56  \\
        Do Not Track                        &1.00 & 0.60 & 0.75  &0.15 & 1.00 & 0.26  &1.00 & 0.60 & 0.75  &1.00 & 0.40 & 0.57  &5   \\
        \hline
        \vspace{-1ex} 
        & \multicolumn{3}{c|}{} 
        & \multicolumn{3}{c|}{} 
        & \multicolumn{3}{c|}{} 
        & \multicolumn{3}{c|}{} &  \\ 
        Micro Average                       &0.83 & 0.54 & 0.66  &0.58 & 0.70 & 0.63  &0.77 & 0.50 & 0.61  &0.72 & 0.64 & {\bf 0.68}  &1115 \\
        Macro Average                       &0.81 & 0.48 & 0.58  &0.50 & 0.70 & 0.55  &0.83 & 0.47 & 0.57  &0.69 & 0.60 & {\bf 0.63}  &1115 \\
        Weighted Average                    &0.79 & 0.54 & 0.62  &0.60 & 0.70 & 0.62  &0.80 & 0.50 & 0.58  &0.71 & 0.64 & {\bf 0.67}  &1115 \\
        \hline
    \end{tabular}  
    \vspace{-3ex}
\end{table*}

As previously discussed, the challenge addressed by our framework and the relevant baselines involves the assignment of appropriate data practice labels to a given paragraph excerpt from a privacy policy, with a selection available from twelve possible categories (classes). Given that a paragraph may contain multiple categories simultaneously, our method is characterised as a multi-class-multi-label classification problem. 
We measure the performance of our framework and others using two types of metrics. To measure classification performance, we use the metrics of precision, recall, and F1 score. To measure the quality of explanations of our method, we use two custom metrics; normalised Levenshtein distance and overlap percentage. 

\subsubsection{Precision, Recall and F1 Score}
\label{sec: PRF1 description}

In multi-class settings, precision (P), recall (R), and F1 Scores are usually reported as micro, macro, and weighted averages. In \textit{micro-averaging}, the average is calculated globally by counting the total true positives and false positives across all classes, whereas in \textit{macro-averaging}, the average of class-wise performance is calculated. In other words, in macro-averaging, each class, including the minority classes, contributes equally to the final number. The \textit{weighted average} is calculated by taking the performance metric of each class, multiplying it by the number of true instances of that class (i.e., the support), and then dividing it by the total number of instances across all classes. This method provides a way to account for the frequency of each class in the dataset when calculating the overall precision. \textit{When we present our results, we include all these averages for completeness and easy comparison with previous work, but we describe the results in terms of macro averages. This is because the macro-average is the most challenging out of these since, to have a higher value, the classifier needs to perform well in minority classes with fewer samples as well.}

\textit{Finally, we also highlight that we consider each annotation and its label as one data point. That is, to obtain a 100\% recall and 100\% precision for a paragraph, a model must get all labels correctly for all the annotations without producing any additional labels or missing any existing labels.} 
\\ \vspace{-2mm}

\subsubsection{Explainability}
\label{sec: quantify explainability}

Our model’s unique approach to producing class labels at the explained classifier stage involves following the idea of one-step chain-of-thought reasoning. This process answers the \emph{explanatory linguistic question of ‘why?’} a particular class label is relevant by identifying the most suitable text chunk from the given paragraph. As the OPP-115 dataset contains annotations by legal experts that justify the assigned labels, we leverage these annotations to evaluate the explainability of our framework. To measure the semantic alignment and overlap between the justifications (reasons) generated by our model and the legal annotations, we employ two metrics: \\ \vspace{-3mm}

\noindent{\textbf{i) Normalised Levenshtein Distance} measures the character-level similarity between two strings. Formally, \emph{Levenshtein Distance} is the minimum number of modifications required to convert one string to another using the operations of insertion, deletion, and substitution. In our case, we use it to measure the distance between the reason produced by our method and the ground-truth legal annotation. We normalise it by dividing it by the maximum length of either the reason or the annotation.} 

\noindent{\textbf{ii) Overlap Percentage} measures the word-level similarity between the reason and annotation. We calculate overlap based on \emph{Jaccard similarity} that evaluates the intersection over the union of words present in the two text pairs.} A perfectly aligned and legal expert-level-like output from our method would have a zero normalised Levenshtein distance and a 100\% overlap with the legal annotation text and vice versa. \\ \vspace{-3mm}

While the language generation-based classification models can be evaluated using the above two metrics, the embedding-based models, such as PrivBERT by design, are black-box and, as such, do not provide explanations for their predictions. To this end, we use the popular framework of LIME~\cite{LIME}.\\ \vspace{-2mm}

\noindent{\textbf{LIME:} Local Interpretable Model-agnostic Explanations 
is a framework we can use to identify `the most important words' that are driving the classification output, and therefore, these LIME words are effectively treated as `explaining' the model prediction. The LIME technique iteratively perturbs the original text sequence and feeds the perturbed versions to the black-box model, observing how these perturbations positively or negatively influence the output prediction class. We can then quantitatively compare those selected LIME words with legal annotations to understand the overlap percentage. However, there are two inherent limitations to using LIME with an embedding-based black box model.
\begin{itemize}[leftmargin=*]
    \item LIME being designed for predictive models, the words it highlights do not always belong to a continuous block in the input text. Therefore, we only consider the word-to-word overlap percentage. The Levenshtein distance, being a character-level operation, is not applicable here.
    \item LIME performs perturbations while observing the effect of the prediction of a single class prediction output's softmax score. Therefore, we can not obtain explanations for a multi-label setting. Instead, we only obtain explanations for the model's most confident class output governed by the highest softmax value over all class labels.     
\end{itemize}
}

%% file: 5_results.tex
\section{Results}
\label{Sec: Results}

\begin{figure*}[t]
    \centering
    \includegraphics[width = 0.83\textwidth]{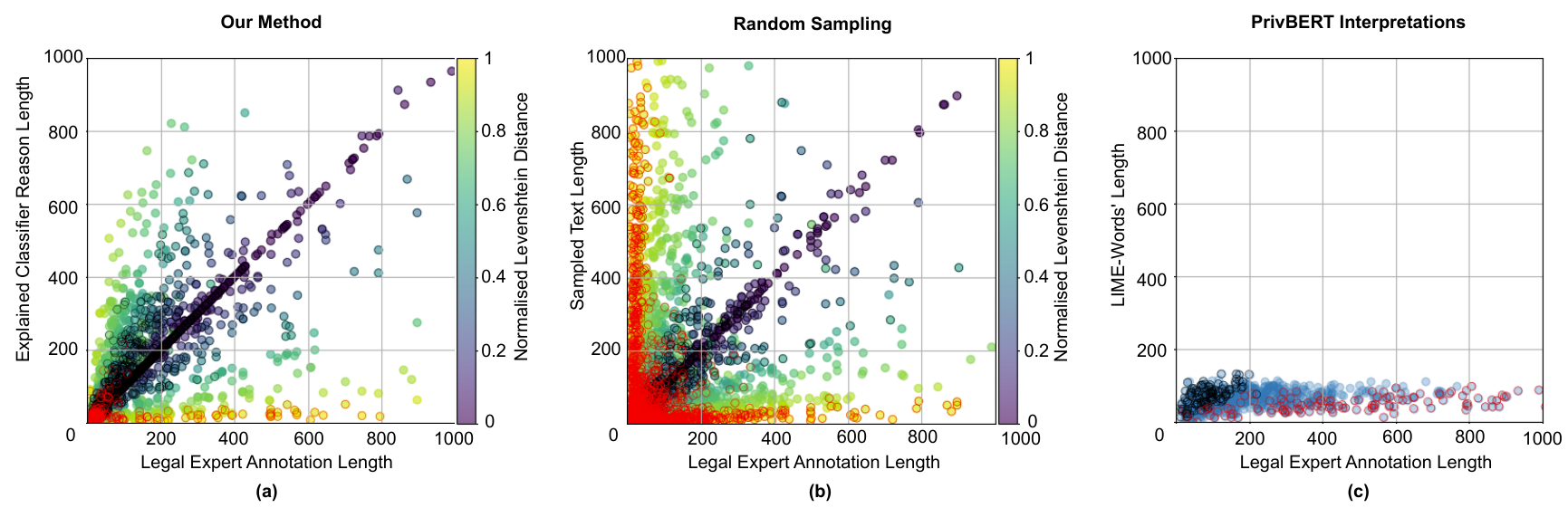} \vspace{-3mm}
    \caption{Explainability visualised: (a): our method's generated, (b): randomly sampled, (c): LIME based PrivBERT interpretations, compared with legal expert annotations. *Subfigure (c) is not colour mapped. **All sub-figures: outlined in red colour are the samples that have less than 10\% overlap with legal expert annotations. Outlined in black are the samples having more than 50\% overlap.}
    \vspace{-2ex} 
    \label{fig:explainability}
\end{figure*}

Next, we present our results, followed by an ablation study to highlight how different components of our overall framework contribute to the final performance. Later, we present the findings on the explainability of predictions.

\subsection{Performance Comparison}

In Tab.~\ref{tab: baselines}, we present class-wise Precision (P), Recall (R), and F1 Scores and their micro, macro, and weighted averages. Although the Table presents results for all metrics, our discussions mainly focus on macro averages because they serve as a representative measure of the overall performance of the models, as discussed in Section IV-E. \\ \vspace{-3mm}

\noindent{\bf Embedding Based Classification:} As can be seen from Tab.~\ref{tab: baselines} (a), among the embedding-based classification methods which are not domain fine-tuned (i.e., GPT2 Embeddings, BERT, and RoBERTa), the macro F1 scores are similar with BERT having the lowest score. This is because BERT performs relatively worse in the absolute minority class \emph{``Do Not Track''}  with only an F1 score of 0.33. The slightly higher performance of RoBERTa over BERT can be attributed to it being pre-trained on a much larger dataset. Also RoBERTa's slightly higher performance over GPT2 embeddings can be attributed to it being an encoder model rather than a decoder model. It is known in the literature that encoder models perform better than decoder models in text classification tasks~\cite{kementchedjhieva2023exploration}. 

PrivBERT, the privacy policy domain adapted model of RoBERTa, performs well with the highest R value of 0.71 while retaining a high P value of 0.83. Therefore, the resulting F1 score of 0.76 indicates that further pre-training has indeed helped for better privacy policy paragraph classification. However, we should also note that it is only a 5.5\% improvement of F1 score compared to RoBERTa. We also observe that it is struggling to recall the class \emph{``practice not covered"}. We also note that our PrivBERT results are lower than those reported by the authors~\cite{privbert}; in this paper, we record the best results we could reproduce with our train-test split. \\ \vspace{-2mm}

\begin{table}[t]
    \centering
    \caption{Ablation Study: All values are macro-average scores} \vspace{-3mm}
    \label{tab:ablations}
    \begin{tabular}{lccc}
        \hline
        \vspace{-1ex} 
        \\
        Description                                & P     & R     & F1  \\        
        \hline 
        \vspace{-1ex} 
        \\
        Explained classifier only                   & 0.38  & 0.85  & 0.48\\
        Explained classifier + entailment verifier  & 0.61  & 0.61  & 0.59\\
        Full pipeline                               & 0.69  & 0.60  & 0.63\\
        \hline
        \vspace{-5ex} 
    \end{tabular}
\end{table}

\noindent{\bf Language Generation Based Classification:} As can be seen from Tab.~\ref{tab: baselines} (b), the macro averaged F1 scores for T5, GPT4, and LlaMA2 are 0.58, 0.55, and 0.57, respectively. Compared to those, our framework has a significantly better performance with a macro-averaged F1 score of 0.63 (i.e., 8.6\%, 14.5\%, and 10.5\% higher than the original results of T5, GPT4, and LLaMA2), indicating the effectiveness of our proposed framework. While our method does not reach the performance levels of embedding-based methods, our method has the advantage of explaining the classification results ({\bf cf.} Sec.~\ref{sec:explainability}). 

Finally, we highlight that we tried to replicate GPT4 zero-shot results outlined in the recent preprint PolicyGPT~\cite{tang2023policygpt} using the same dataset. Despite multiple attempts and communications with the authors, we could not reproduce the results presented in that paper. We believe that authors may have had some pre-filtering of data and different evaluation metrics (e.g., considering a classification as successful even if one predicted label is true among multiple annotation labels per paragraph - in contrast, our evaluation is more rigorous as explained in Sec.~\ref{sec: PRF1 description}), making the exact problem they address different from ours and other comparable baselines, such as PrivBERT. 

\subsection{Ablation Study}

To provide further evidence on the overall effectiveness of our framework and how contributions from individual components of our framework work together, we conducted an ablation study. That is, we evaluate the performance of our framework by progressively adding modules starting from the explained classifier. We show the results in Tab.~\ref{tab:ablations}. 

The explained classifier could be considered a `thought generator' where, for each paragraph, it tries to generate multiple `class output and reason' pairs until the token generation limit is exhausted. In that process, it recalls many of the correct pairs (0.85); however, some of them are not accurate, as indicated by the low precision of 0.38. As soon as we train an entailment verifier to filter out incorrect outputs, our precision improves to 0.61. Nonetheless, it decreases the recall because some correct classifications are filtered, specifically those about which the entailment verifier is not confident. As we employ the full pipeline, including blank filler, we obtain the highest macro-averaged precision of 0.69. Finally, we point out that even without the blank filler, our model's macro-averaged F1 score is higher than zero-shot GPT4 and LLaMA2. 

\subsection{Explainability}
\label{sec:explainability}

Next, we compare the explainability provided by our method (best-performing language generation-based method) and PrivBERT (best-performing embedding-based method) using the metrics described in Sec.~\ref{sec: quantify explainability}. More specifically, we use normalised Levenshtein distance and overlap percentage for our method and LIME-based overlap percentage for PrivBERT. We report the results for our held-out test set.

\begin{table}[t]
    \centering
    \caption{Overlap Percentage} \vspace{-2mm}
    \label{tab:overlaps}
    \begin{tabular}{rrrr}
        \hline
        \vspace{-1ex} 
        \\
        Overlap (\%)  & Our Method & Random sampling & PrivBERT \\
        \hline
        \vspace{-1ex} 
        \\
        50 - 100        & \textbf{57.9} & 16.2 & 18.3\\
        10 - 50         & \textbf{33.3} & 37.9 & 62.9\\
        less than 10    & \textbf{8.8}  & 45.9 & 18.8\\
        \hline
        \vspace{-1ex} 
        \\
        \multicolumn{4}{c}{*57.9 indicates that 57.9\% samples of our method's predictions } \\
        \multicolumn{4}{c}{overlap with legal-expert annotations by 50 to 100\%. }\\
        \vspace{-5ex} 
    \end{tabular}
    \vspace{-3mm}
\end{table}

First, we present Levenshtein distance results in Fig.~\ref{fig:explainability} (a) and (b). Each scatter dot represents a data point in our test set. The x-axis represents the legal annotation's character length, while the y-axis represents the generated reason's character length. Each data point is coloured according to the normalised Levenshtein distance between the two texts. A diagonal datapoint with 0 distance indicates a perfect prediction similar to \emph{`what a legal expert would have annotated'}.

We show the results of our method in Fig.~\ref{fig:explainability} (a). For comparison, in Fig.~\ref{fig:explainability} (b) we present the same result for a random text generation baseline. That is, we run a separate experiment where, for each prediction, we randomly sample a text from the same paragraph and assume it as the reason generated by the model. This random sampling is done according to the annotation length to paragraph length ratio distribution of the training dataset to mimic a realistic sampling process. 

We observe from the results that the generated reason distribution of our method is more positively correlated with the legal expert annotation distribution, unlike a randomly selected sample (from the same paragraph) distribution. In our case, most of the points are around or in the direction of the diagonal. In contrast, in the random case, there are more samples spread near the x-axis and y-axis, indicating significant differences between the two texts. In the figure, we outline all the samples with little or no overlap with the legal expert annotation despite some having a low normalised Levenshtein distance between them, in red. As can be seen, our method has significantly fewer such points. We further quantitatively analyse overlap percentages later in this section.

Observing the length of the PrivBERT's LIME-words to legal expert annotation length distribution in Fig.~\ref{fig:explainability} (c), we can visually identify some drawbacks with embedding models. First, LIME can only interpret the explainability of PrivBERT's most confident output; therefore, the number of samples we can analyse is lower. Next, LIME being designed for predictive models, the words it identifies may not necessarily belong to a continuous block, and it can only consider a certain number of perturbations in a selected paragraph. Therefore, the samples are distributed more along the direction of the x-axis with LIME-words' length capped at $\sim$150 characters. As we do not consider normalised Levenshtein distance in this figure, darker shades of blue only represent densely packed sample points. Outlined in red represents the same meaning as with subfigures (a) and (b). 

To quantitatively analyse the explainability, in Tab.~\ref{tab:overlaps}, we present the results for overlap percentages. We observe that with our method, 57.9\% of the predictions have at least 50\% overlap with legal annotations. However, in contrast to that, nearly 45.9\% of randomly selected text had less than 10\% overlap with the legally annotated text (these samples with less than 10\% overlap are outlined in red colour in all sub-figures in Fig.~\ref{fig:explainability}). These results show that our method's explainability pre-dominantly overlaps with legal annotations and quantitatively, there is at least a 10\% overlap for more than 90\% of data samples with our method. 

When we consider LIME-word based overlap percentage for PrivBERT, only around 18.3\% of LIME words overlap 50\% or more with the legal annotations and even for random sampling, this overlap count was 16.2\%. Also, a similar percentage (18.8\%) of samples had less than 10\% of overlap. From qualitative observations, we further identified that most LIME words concentrate with class-specific words such as ``third" for \emph{``third party sharing/collection"}. This concludes that even when looking at the most confident output of PrivBERT, its quantified explainability is really low compared to our method.

%% file: 6_conclusion.tex
\section{Conclusion}
\label{sec:conclusion}

We proposed an entailment-driven LLM-based framework for privacy policy paragraph classification and for providing explanations behind those predictions. Our training pipeline consists of an explained classifier, blank filler, and an entailment verifier that outperformed other language generation-based baselines such as T5, GPT4, and LLaMA2 by $\sim$8\%--14\%. 
The key reason for this is that our framework, inspired by one-step Chain of Thoughts (CoTs) reasoning, avoids the commonplace hallucination problem of LLMs by providing a reason for each classification label. Using the blank filler that re-predicts the same reasoning that subsequently undergoes the entailment verification process, our method can filter such hallucinated outcomes effectively. As a result, our model has a macro-average precision increase of 38\% compared to GPT4. Though the proposed framework does not achieve the performance levels of embedding-based models such as PrivBERT, it provides explanations behind the label predictions, which is useful in the context of privacy and usability. To this end, we showed that our method generates reasoning texts that are likely to be at least 50\% or more overlapping with what a legal expert would have reasoned. Overall, our results show that while LLMs can be useful for providing more user-friendlier means to access privacy policies, they are not that useful in their vanilla form. Rather, it is necessary to have auxiliary steps as we proposed in our framework.